\documentclass[conference]{IEEEtran}
\IEEEoverridecommandlockouts
\usepackage{cite}
\usepackage{amsmath,amssymb,amsfonts}
\usepackage{algorithmic}
\usepackage{graphicx}
\usepackage{textcomp}
\usepackage{xcolor}
\usepackage{balance}
\usepackage{flushend}
\usepackage{placeins}
\def\BibTeX{{\rm B\kern-.05em{\sc i\kern-.025em b}\kern-.08em
    T\kern-.1667em\lower.7ex\hbox{E}\kern-.125emX}}

\usepackage{tikz}
\usepackage[hyphens]{url} 
\usepackage{pgfplots}
\usepackage[hidelinks]{hyperref}  

\usepackage{booktabs} 

\usepackage{lipsum}
    
\newcommand{\term}[1]{\textit{#1}}
\newcommand{\comment}[1]{}

\pgfplotsset{select coords between index/.style 2 args={
    x filter/.code={
        \ifnum\coordindex<#1\fi
        \ifnum\coordindex>#2\fi
    }
}}

\begin{document}

%
%
%
%
\title{Distributed LSTM-Learning from Differentially Private Label Proportions\\
\thanks{
This research has been funded by the Federal Ministry for Economic Affairs and Climate Action of Germany under grant no. 19S21005N GAIA-X 4 ROMS, as well as the Federal Ministry of Education and Research of Germany and the state of North-Rhine Westphalia as part of the Lamarr-Institute for Machine Learning and Artificial Intelligence, LAMARR22B.
}
}


\author{
\IEEEauthorblockN{Timon Sachweh}
\IEEEauthorblockA{
\textit{TU Dortmund University}\\
Dortmund, Germany \\
0000-0002-0347-5760}
\and
\IEEEauthorblockN{Daniel Boiar}
\IEEEauthorblockA{
\textit{TU Dortmund University}\\
Dortmund, Germany \\
0000-0002-6856-9848}
\and
\IEEEauthorblockN{Thomas Liebig}
\IEEEauthorblockA{
\textit{TU Dortmund University}\\
Dortmund, Germany \\
0000-0002-9841-1101}
}
\maketitle

\begin{abstract}
Data privacy and decentralised data collection has become more and more popular in recent years.
In order to solve issues with privacy, communication bandwidth and learning from spatio-temporal data, we will propose two efficient models which use Differential Privacy and decentralized LSTM-Learning:
One, in which a Long Short Term Memory (LSTM) model is learned for extracting local temporal node constraints and feeding them into a Dense-Layer (LabelProportionToLocal).
The other approach extends the first one by fetching histogram data from the neighbors and joining the information with the LSTM output (LabelProportionToDense).
For evaluation two popular datasets are used: Pems-Bay and METR-LA.
Additionally, we provide an own dataset, which is based on LuST.
The evaluation will show the tradeoff between performance and data privacy.
\end{abstract}


\section{Introduction}

In the last few years, the increased popularity of the Internet of Things (IoT) has led to 
an increasing amount of decentralized data collection. 
According to the current state, the data is usually sent to a central instance, where the data is processed.
Centralized learning methods have several problems:

The first 
aspect that becomes clear is the lack of \term{data protection}.
Especially with the introduction of the \term{General Data Protection Regulation (GDPR)} \cite{Mondschein2019}, a lot has changed in terms of \term{data privacy}, which must be implemented by everyone using sensitive data.
Because of divergent goals between data protection and learning from data, this is an urgent topic.


Another aspect is the ever-increasing number of Internet participants.
Taking into account that the maximum Internet traffic capacity is not increasing at the same rate, the bandwidth per device is shrinking.
In the long term, this can result in bottlenecks for participants that need high bandwidth rates.

The two proposed \term{fully distributed deep learning} algorithms will ensure flexible \term{data privacy} by setting a hyperparameter to balance both aspects: privacy and prediction accuracy.

\subsection*{Existing Approaches}

Since there are already algorithms that cover parts of the challenges of traffic flow prediction with privacy aspects, we will briefly distinguish our approach from the existing ones.
%
\newcommand{\tableHead}[1]{\textbf{#1}}

Most algorithms usually focus on one of the two relevant properties, \term{privacy} or \term{prediction accuracy}.
For example, the \term{dp-LLP} algorithm \cite{Sachweh2021:llp}, introduced by Sachweh~et~al., is a fully distributed learning algorithm that uses only locally collected data, as well as data from the neighbor nodes, to predict traffic flow.
The authors introduced a variant of \term{Label Proportions}, originally developed in \cite{4470249, stolpe2011learning, Liebig2015:labelproportions}, extended by \term{Differential Privacy} to ensure that data transfer is protected.
Key advantages of this approach are \term{less data traffic} during execution and full \term{Data Privacy}.
One negative aspect is the low complexity of the integrated \term{$k$-Means} learning algorithm, which results in worse prediction accuracy.
In addition, the authors do not use time-dependent features, which is an important feature space especially for traffic flow.

Another algorithm that uses \term{Label Proportions} for data transfer was developed by Dulac~et~al. \cite{DulacArnold2019:DeepMultiClassLLP}.
The authors use the original \term{Learning from Label Proportions (LLP)} \cite{stolpe2011learning} approach and change the learning model by using a \term{Long-Short Term Memory (LSTM)} model.
This more complex model results in higher prediction accuracy results with lower energy consumption, shown using the MNIST dataset.
Since this approach lacks time-dependent features too, it is not perfectly suitable for vehicle traffic prediction. 

In contrast, \cite{Yu2018:SpatioTempGCNNTraffic} makes full use of spatio-temporal data. 
One global \term{Graph Convolutional Neural Net (GraphCNN)} model is learned, with decentralized sensors as nodes in the graph.
Local sensor data trains the local neighborhood in the graph.
With this approach original data labels are propagated.
Therefore it is not \term{privacy preserving} at all.
This approach also has high \term{energy consumption} and is not ready for \term{peer to peer} scenarios.

Our approach builds on the experience of the papers presented, and will leverage the data protection properties from \cite{Sachweh2021:llp}, as well as comparable performance to \cite{DulacArnold2019:DeepMultiClassLLP,Yu2018:SpatioTempGCNNTraffic}.

\section{Fundamental Work}


In the following we provide basic knowledge of various privacy-preserving data transfer 
methods, as well as an explanation of \term{Differential Privacy}.
\term{Long-Short Term Memory (LSTM)} models and the \term{$k$ Nearest Neighbor ($k$NN)} classifier will be described as well.

\subsection{Privacy Preserving Data Transfer}
The distributed learning setting has led to more and more private data transfer methods.
Concerning data transfer the following general concepts can ensure privacy:
\begin{itemize}
    \item \term{Homomorphic Encryption}\cite{rivest1978method} ensures that encrypted data is transformed into another space with similar learnable features. Therefore models can be learned with encrypted data, as shown in \cite{acar2018survey}. One downside of this method is the high usage of computation resources.
	\item \term{Masking} addresses this problem by inserting \term{Camouflage Values}. Those are fake values, that mask the original data points.
\item An alternative to \term{Masking} is data aggregation. This has the advantage of data compression and a lot of variety in search queries. Temporal aggregation of labels in so-called label proportions is the core idea of related works \cite{stolpe2011learning,Liebig2015:labelproportions,Sachweh2021:llp,DulacArnold2019:DeepMultiClassLLP}. Besides the possibility to hide individual data points in an aggregate, \term{Learning from Label Proportions (LLP)} also reduces communication costs and energy consumption.
\end{itemize}

Because of the lack of a central authority in a fully distributed scenario, encryption does not directly provide a solution. 
Recent work that combines learning from label proportions with differential privacy (compare next section) \cite{Sachweh2021:llp} is promising, and we therefore utilize \term{Data Aggregation} for our approach aswell.

\subsection{Differential Privacy}
\label{subsec:differential-privacy}

\term{Data Aggregation} methods, which are applied on very pure data are problematic, because resulting histograms contain all information, although data points were aggregated.

Therefore \cite{Sachweh2021:llp} extended \term{Data Aggregation} (building histograms) by adding \term{Differential Privacy} to the histograms.
\term{Differential Privacy} adds noise to each bin to ensure a specific privacy guarantee.

\subsubsection{Differential Privacy Definition}

In general, an algorithm is $(\varepsilon , \delta )$-differentially private, if for all $S \subseteq R$ \autoref{eq:differential-privacy} is valid \cite{dwork2008differential}. $M: D \rightarrow R$ denotes a randomized algorithm and $D',D'' \in D$ are sets, which differ at most by one element ($|| D' - D'' ||_1 \leq 1$).

\begin{align}
    Pr[M(D') \in S] \leq e^{\varepsilon} Pr[M(D'') \in S] + \delta
    \label{eq:differential-privacy}
\end{align}

The definition states that the probability distributions, that the output of $M$ is in $S$ for different inputs $D'$ and $D''$ differ at most by a factor of $e^\varepsilon$ and a constant value of $\delta$.
In our approach the constant factor $\delta$ is $0$ and called $\varepsilon$-differentially private if it satisfies the \autoref{eq:differential-privacy}.

\subsubsection{Sensitivity}

Sensitivity is needed to gain information about the maximum influence of a single data point in a dataset $D$.
Therefore $l_1$-sensitivity is defined as follows:

\begin{align}
    \Delta f = max_{\substack{D', D'' \in D, \\ ||D'-D''||_1=1}} ||M(D') - M(D'')||_1
    \label{eq:sensitifity}
\end{align}

It states that every two subsets $D', D'' \in D$, which differ exactly by one element, are checked for the maximum $l_1$ distance of the outputs of $M$ with different inputs $D'$ and $D''$.

\subsubsection{Laplacian Noise}

To satisfy \autoref{eq:differential-privacy}, we need to apply noise to each bin.
The noise calculation must be scaled by the privacy parameter $\epsilon$, as well as the \term{$l_1$-sensitivity}.
To achieve this, we introduce \term{Laplacian distribution}, defined as follows:

\begin{align}
    lap(x | \sigma, \mu) &= \frac{1}{2 \sigma} e^{-\frac{|x-\mu|}{\sigma}} \\
    \label{eq:laplace}
\end{align}

Parameter $\mu$ sets the mean value. 
In our case, the mean is $0$.
By inserting $\frac{\Delta f}{\varepsilon}$ for $\sigma$, the variance of \term{Laplacian Distribution} is dependent on the \term{$l_1$-sensitivity} and the privacy factor $\epsilon$.
Theorem 3.6 in \cite{dwork2014algorithmic} proves, that the \term{Laplace distribution} ensures the $(\varepsilon, 0)$-Differential Privacy border for $\sigma=\frac{\Delta f}{\varepsilon}$ and $\mu = 0$.

By varying $\varepsilon$, the privacy guarantee can be changed.
Experiments have shown 
$\epsilon = 0.1$ is a good setting in order to ensure privacy, but keeps enough information for learning on the \term{differentially private} data.

\subsection{Long-Short Term Memory (LSTM)}

\label{subsec:model-architecture}
Long short-term memory (LSTM) \cite{hochreiter1997long} is a model often used for time series prediction (e.g. traffic \cite{huang2020lsgcn}). Though temporal loops can enrich Feedforward Networks to process sequential data in so-called Recurrent Neural networks\cite{manolios1994first}, this solely Markovian modeling approach has drawbacks. 
The concept of recurrent neural networks is to process a sequence of information, but these models are not capable of considering different inputs and outputs together. Even if the information is connected, it was considered individually. This poses various challenges for many tasks. Clearly, one needs to know the succeeding data to predict the future since the two are connected.
 In some sense, recurrent neural networks serve as a memory that collects and stores information about what the system has computed so far. A recurrent neural network system can look back at some steps to use previous information for current knowledge.

In LSTM, the data transfer process is the same as in standard recurrent neural networks. However, the operation to propagate the information differs. As the information passes through, the model selects which information to process further and which to let pass. The network structure consists of cells, each consisting of 3 gates (input, output, forget). Each of the gates themselves can be considered a feedforward network. However, they are connected by the state of the cell. The state of a cell acts as a path to transmit information. Cells are, therefore, memories.

\subsection{$k$ Nearest Neighbor ($k$NN)}
\label{sec:knn}
The \term{$k$ Nearest Neighbor ($k$NN)} classifier is a non-parametric supervised learning algorithm.
It can be adopted to be well suited for traffic prediction \cite{yang2019k}.
Originally it was developed by Fix and Hodges in 1951 \cite{knnFirstApproach}.

The general concept is to store all measured data.
For example, the traffic flow rate and corresponding time as one feature vector.
There is no training phase.
The prediction is made by determining the distance to all saved feature vectors and getting the labels of the closest $k$.
Finally prediction is made by a majority voting over the closest $k$ labels.

\section{Our Approach}
\label{sec:approach}

Our approach is combining the advantages of DP-LLP \cite{Sachweh2021:llp}, and \cite{DulacArnold2019:DeepMultiClassLLP} to build a \term{fully decentralized} learning algorithm, which ensures \term{privacy} and results in good forecasting performance.
First, we will describe the general data exchange between the distributed nodes and the integration of \term{Differential Privacy}.
Afterwards, we show the advanced \term{LSTM} model, which uses differentially private neighbor information to learn.

\subsection{Distributed Network}

\begin{figure}
    \centering
    \begin{tikzpicture}[%
    graphnode/.pic={
		\draw (0,0) circle (0.5);
		\begin{axis}[
			width=2.4cm,
			height=2.3cm,
			axis line style={draw=none},
      		tick style={draw=none},
      		xtick={3},
      		ytick={3},
      		at={(-125,-65)}
		]
			\addplot[smooth] coordinates{(-1.2,-0.1) (-1,0.2) (-0.8,0)(-0.4,0.6) (-0.2,-0.5) (0,0.4) (0.2,-0.2) (0.4,0.3) (0.7,-0.1)  (0.9,0.3)};	
		\end{axis}
     },
     histogram/.pic={
     	\draw[fill,lightgray] (0,0) rectangle +(0.1,0.2);
     	\draw[fill,gray] (0.1,0) rectangle +(0.1,0.3);
     	\draw[fill,black] (0.2,0) rectangle +(0.1,0.1);
     }]
	\draw (0,0) pic {graphnode};
	\node (j) at (0,0-0.8) {$j$};
	
	\draw (1.5,1.5) pic[gray] {graphnode};
	\node (aj) at (1.5+0.5,1.5+0.7) {$n_2(j)$};
	
	\draw (-1.5,1.6) pic[gray] {graphnode};
	\node (bj) at (-1.5-0.5,1.6+0.7) {$n_1(j)$};
	
	\draw (-2.8,-0.5) pic[gray] {graphnode};
	
	\draw (2,-0.5) pic[gray] {graphnode};
	\node (dj) at (2+0.5,-0.5-0.7) {$n_3(j)$};
	
	\draw[<->] (0.4,0.4) -- (1.1,1.1);
	\draw (0.8,0.5) pic {histogram};
	
	\draw[<->] (0.5,-0.2) -- (1.45,-0.4);
	\draw (-1.1,0.5) pic {histogram};
	
	
	\draw[<->] (-0.4,0.4) -- (-1.1,1.2);
	\draw (0.8,-0.7) pic {histogram};
	
	\draw[gray, dashed] (-2.5,0) -- (-1.9, 1.2);
	\draw[gray, dashed] (-2.4,-0.8) -- (-1.3, -1.3);
	\draw[gray, dashed] (-2.9,0.05) -- (-3.3,1);
	\draw[gray, dashed] (0.9,1.5) -- (-1, 1.6);
	\draw[gray, dashed] (1.7,0.95) -- (1.9,0.05);

\end{tikzpicture}
    \caption{Network Architecture: Connected nodes specified by an adjacent matrix. Node $j$ is the observed node and $n_1(j), n_2(j), n_3(j)$ are the direct neighbor nodes, that are sending aggregated histograms to $j$ for learning.}
    \label{fig:graphSetup}
\end{figure}
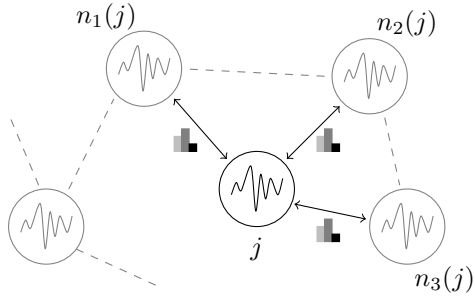

A general distributed network setup is shown in \autoref{fig:graphSetup}.
We have a list of nodes and an adjacent matrix, which defines the edges between nodes.

We denote $j$ the node that is currently observed.
For all $i \in |N_j|$, $n_i(j)$ are the neighbors of $j$ and $|N_j|$ denotes the count of neighbors from $j$, shown in \autoref{fig:graphSetup}.
In our approach, data is only transferred between direct neighbors, illustrated with arrows between $j$ and its neighbors.
The histograms in \autoref{fig:graphSetup} denote that no original data is sent via the network but just aggregated intervals over larger time frames, so-called \term{buckets}.
All transferred data ensure $\epsilon$-Differential Privacy and are used by the neighbors to learn a \term{LSTM} model.

\subsection{Distributed Data Exchange}

As mentioned before, transferred data must by $\epsilon$-differentially private. 
To ensure this, data is first discretized and then sliced in fixed time windows of size $w$.
Each window contains $x_l$ for $l \in [0, w]$ data points.
For each window a histogram is calculated to aggregate the data.
Additionally laplacian noise is added with $\sigma = \frac{1}{\varepsilon}$ and $\mu = 0$.
The sensitivity is fixed to $1$ because the maximum influence of a single data point in a counting query is $1$.
Finally we get an $\epsilon$-differentially private histogram, which is send to the neighbors.

However, to say that the whole algorithm will be $\varepsilon$-differentially private, one has to show that later processing on $\varepsilon$-differentially private data is at least $\varepsilon$-differentially private.
Fortunately, Proposition 2.1 from \cite{dwork2014algorithmic} proves exactly this.
The authors show, that every \term{Post-Processing} on data that is $(\varepsilon, \delta)$-differentially private will result in a $(\varepsilon, \delta)$-differentially private outcome, too.

Using this Proposition, we can prove that applying noise at each transferred histogram is sufficient to ensure a fully $\varepsilon$-differentially private algorithm.

\subsection{Model Architecture}

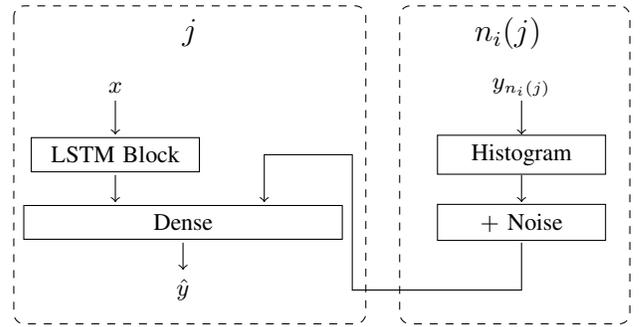
\begin{figure}[!h]
    \centering
    \begin{tikzpicture}[scale=0.9]
    	\tikzstyle{every node}=[font=\small]
    	\node (titleJ) at (1.1,0.8) {\large$j$};
    	\node (titleNeighbor) at (5.8,0.8) {\large$n_i(j)$};
    
    	\node (x) at (0,0) {$x$};
    	\node[draw, text width=2cm, align=center] (lstm) at (0,-1) {LSTM Block};
    	\node[draw, text width=4cm, align=center] (dense) at (1,-2) {Dense};
    	\node[] (yhat) at (1,-3) {$\hat{y}$};
    	
    	\node (yneighbor) at (6,0) {$y_{n_i(j)}$};
    	\node[draw, text width=2cm, align=center] (hist) at (6,-1) {Histogram};
    	\node[draw, text width=2cm, align=center] (noise) at (6,-2) {$+$ Noise};
    	
    	\draw[->] (0,-0.2) -- (0,-0.7);
    	\draw[->] (0,-1.3) -- (0,-1.7);
    	\draw[->] (1,-2.3) -- (1,-2.7);
    	
    	\draw[->] (6,-0.2) -- (6,-0.7);
    	\draw[->] (6,-1.3) -- (6,-1.7);
    	
    	\draw[->] (6,-2.3) -- (6,-3) -- (3.5,-3) -- (3.5,-1) -- (2.2,-1) -- (2.2,-1.7);
    	
    	\draw[dashed,rounded corners] (-1.5,1.2) rectangle (3.7,-3.5);
    	\draw[dashed,rounded corners] (4.2,1.2) rectangle (7.6,-3.5);
    \end{tikzpicture}
    \caption{Proposed distributed Label Proportion LSTM architecture. The LSTM Block contains an LSTM layer, followed by a ReLU and a local linear layer.}
    \label{fig:archConcat}
\end{figure}

In principle, our model can be split into two parts as visualized in \autoref{fig:archConcat}.
The first part comprises the node-wise learning block, where each node uses its data for training.
The second block explicitly trains the last layer of the model by using aggregated data from the neighbors as additional information.

In \autoref{fig:archConcat} it can be seen that we are using the local data of node $j$, denoted as $x$, as inputs for the \term{LSTM} model.
Parallelly, the aggregated spatial data from neighbors is built up and sent to node $j$.
Node $j$ then uses the aggregated data to improve the learning of the network's last \term{Dense-Layer}.
Finally, the outcome of the \term{Dense-Layer} is the prediction $\hat{y}$, from which the gradient could be determined, and all weights will be updated.

In the following, we will describe in more detail how the learning in both phases works.

\subsubsection{Local Node Learning}
\label{subsubsec:node-learning}

\begin{figure}[!h]
    \centering
    \begin{tikzpicture}[scale=4]
	\draw[] (-1.4,0) rectangle +(0.3,0.1) node[pos=.5] {$x_{speed}$};
	\draw[] (-1.1,0) rectangle +(0.1,0.1) node[pos=.5] {1};
	\draw[] (-1,0) rectangle +(0.1,0.1) node[pos=.5] {0};
	\draw[] (-0.9,0) rectangle +(0.1,0.1) node[pos=.5] {1};
	\draw[] (-0.8,0) rectangle +(0.1,0.1) node[pos=.5] {1};
	\draw[] (-0.7,0) rectangle +(0.1,0.1) node[pos=.5] {2};
	\draw[] (-0.6,0) rectangle +(0.1,0.1) node[pos=.5] {0};
	
	\draw[] (-1.4,-0.11) rectangle +(0.3,0.1) node[pos=.5] {$x_{density}$};
	\draw[] (-1.1,-0.11) rectangle +(0.1,0.1) node[pos=.5] {2};
	\draw[] (-1,-0.11) rectangle +(0.1,0.1) node[pos=.5] {0};
	\draw[] (-0.9,-0.11) rectangle +(0.1,0.1) node[pos=.5] {2};
	\draw[] (-0.8,-0.11) rectangle +(0.1,0.1) node[pos=.5] {1};
	\draw[] (-0.7,-0.11) rectangle +(0.1,0.1) node[pos=.5] {1};
	\draw[] (-0.6,-0.11) rectangle +(0.1,0.1) node[pos=.5] {0};
	
	\node (measurementText) at (-0.94,-0.16) {LSTM + RELU outputs};
	
	\draw[rounded corners, draw=black, fill,lightgray, opacity=0.7] (-1.09,-0.105) rectangle +(0.08,0.2) node[pos=.5] {};
	\draw[rounded corners, draw=black, fill,gray, opacity=0.7] (-0.99,-0.105) rectangle +(0.08,0.2) node[pos=.5] {};
	
	\draw[-] (-1.05,0.1) -- (-1.05,0.2);
	\draw[->] (-1.05,0.2) -- (-0.25,0.2);
	
	\draw[-] (-0.95,0.15) -- (-0.95,0.1);
	\draw[-] (-0.95,0.15) -- (-0.4,0.15);
	\draw[-] (-0.4,-0.1) -- (-0.4,0.15);
	\draw[->] (-0.4,-0.1) -- (-0.25,-0.1);

    \node (matrix1Text) at (0.05,0.2) {
    $\begin{pmatrix} 
	1 & 0 \\
	0 & 1\\
	\end{pmatrix}
	+ b $
    };
    \node (matrix2Text) at (0.05,-0.1) {
    $\begin{pmatrix} 
	1 & 0 \\
	0 & 1\\
	\end{pmatrix}
	+ b $
    };
    \node (labelJ) at (-0.94,0.32) {$j$};
    \draw[dashed,rounded corners] (-1.6,-0.25) rectangle +(1.9,0.65);
\end{tikzpicture}
    \caption{Local Linear Layer. The Parameters of the matrices are not shared over time stamps, but are trainable and initialized with the identity matrix.}
    \label{fig:localLinearLayer}
\end{figure}
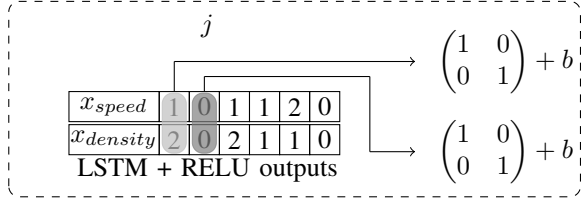

The local node learning phase consists of a \term{LSTM} \cite{Hochreiter97:LSTM, Guo:IMV2019}, which builds features containing temporal dependencies.
Features are learned on fixed time windows of $w = 12$ time steps (around $1$ hour in PemsBay).
By replacing $k-Means$ with a \term{LSTM} model, we gain more learned information about time-dependent features.
When using multiple features, e.g., speed and density, they are handled independently by the \term{LSTM}.
The output is then limited to the $R_+$ value range by a downstream \term{Rectified Linear Unit (ReLU)} activation function.
Outputs of the \term{ReLU} are used to feed the \term{Local Linear Layer}, which is initialized by the \term{identity matrix}.
The reason for this is first to let all information through and later, during the training, modify the weights to gain more information.
The process of feeding outputs of the \term{ReLU} into the initialized \term{Local Linear Layer} is denoted in \autoref{fig:localLinearLayer}.

The first block of the model ends before feeding information to the \term{Local Linear Layer}.
This first phase uses the local information for learning. However, because both blocks of the model are updated in a single \term{Backpropagation}, we need to describe the second part before determining the loss calculation.

\subsubsection{Learning with Aggregated Spatial Information}

In the second phase, we are using the aggregated spatial neighbor information.
To enable this, we use transferred noisy histograms as described previously.
Thus for each time frame we have $|N|$ histograms of $N$ different direct neighbors.

These histograms are also used as inputs for the model -
but only for the \term{Dense-Layer}.
To fix the length of the additional input vector, we had to average the received histograms.
This is needed because we cannot guarantee that the number of neighbors is always the same.

The resulting averaged histogram is concatenated with the outputs of the \term{LSTM + ReLU + Local Linear Layer} and fed into the \term{Dense-Layer} as inputs. The Dense-Layer is applied channelwise, such that the histograms of speed and density are concatenated with time features of speed and density individually.
The arrows denote this data flow from the \term{LSTM} and from the neighbor histogram in \autoref{fig:archConcat}.

The main advantage of this approach is to use more information from which the model can be trained.
Therefore we have updated weights with information from neighboring nodes, which eventually result in better prediction behaviors after the \term{Dense-Layer}.

Nevertheless, there are also some downsides to using this approach.
For example, when predicting a specific node, one must gather the neighbors' histogram information.
Therefore, a stable network connection is required.
To work around this problem, one can use previous neighbors' histogram values or the own histograms of node $j$.
We consider this architecture not dependent on the loss regularization term only and assume it is more accurate because of additional learned information.
A similar assumption was also made by the authors from \cite{stolpe2011learning,Sachweh2021:llp}.

\subsubsection{Loss Calculation}

To this point, we described the data flow from input $x$ and the histograms to the prediction $\hat{y}$.
For development, \term{PyTorch} \cite{NEURIPS2019_9015}, \term{PyTorch Geometric} \cite{Rozemberczki2021:TorchGeoTemporal} and a modified Pytorch implementation of \cite{Guo:IMV2019} were used.
Therefore, all layers in the model architecture result in one \term{Gradient Graph}.
Based on this graph, the loss can be backpropagated.
We used the \term{Mean Squared Error (MSE)} loss for calculating the deviation between prediction and actual data values:

\begin{align}
    loss_{MSE} = \frac{1}{|B|} \sum_{i}^{|B|} (\hat{y}_i - y_i)^2
\end{align}

Because we are using batches with size $|B|$, the squared error loss is calculated for every prediction in the batch, and afterwards, the total amount is normalized by $|B|$.
With this setup, we can propagate the error back to both, the \term{Dense-Layer} with histograms from the neighbors and the \term{LSTM} with inputs of $x$.

\section{Experimental Evaluation}

The experimental results compare our approach against the state-of-the-art \term{kNN} in a centralized computation setting (compare \autoref{sec:knn}).
Performance results are measured on the real-world, large-scale traffic datasets \term{Pems-Bay} \cite{li2018dcrnn_traffic}, \term{METR-LA} \cite{jagadish2014big} and an own generated dataset from \term{Luxembourg SUMO Traffic (LuST)} scenario \cite{codeca2017:luxembourg}. 
We aim to show the level of privacy that we can reach with our distributed learning approach by putting it in relation to the prediction accuracy. 
Before we go into the experiment settings and results in detail, we will briefly describe the datasets and introduce relevant steps in the evaluation.

\subsection{Datasets}

First we will give a detailed overview over the used datasets, especially how the \term{LuST} dataset is built up.

\subsubsection{Pems-Bay}

The \term{Pems-Bay} dataset was collected by the California Transportation Agencies (CalTrans) utilizing the \term{Performance Measurement System (PeMS)}. 
The dataset is based on 325 Bay Area sensors collected from Jan 1st, 2017, to May 31st, 2017, in 5-minute intervals. 
Each data point contains information about the traffic density and a normalized time value.
\term{Pems-Bay} is mainly used to verify the prediction accuracy applicable to non-euclidean structural models.

\subsubsection{METR-LA}

\term{METR-LA} is a similar dataset as \term{Pems-Bay}. 
The containing data was collected from \term{Los Angeles County Highway} ring detectors from March 1st, 2012, to June 30st, 2012.
In total, 207 sensors were used to collect traffic density data.

\subsubsection{LuST}

As the third dataset, we introduce a new one based on the \term{Luxembourg SUMO Traffic (LuST)} \cite{7906642} scenario.
This scenario is executed in the \term{Simulation of Urban Mobility (SUMO)} environment \cite{Lopez2018:sumo}, which was built in order to have a stable basis to develop and test data based mobility solutions.

We let the data simulation run and extracted the traffic counts for every 5 minutes of each street and the corresponding speeds.
As metadata, we collected the graph information and built an adjacency matrix in which the relations between streets were stored.
The resulting dataset contains traffic density and speed for the Luxembourg simulation covering an entire day.

The big advantage of this dataset generation is, that the same process can be executed on a different simulation, or for longer intervals.
In our setting we use the short time frame of one day for analyzing, whether the model can learn on short time periods too.

Our prepared \term{LuST} dataset can be downloaded from the following google drive: \url{https://drive.google.com/uc?export=download&id=1OjPkvptYb22eThm-0zOArBEFA_aLifqp}.
Alternatively one can use the implemented \term{LuSTDatasetLoader} for loading the dataset with PyTorch.

\subsection{Metric}

All datasets are designed to predict a forecast value.
For example, \term{Pems-Bay} and \term{LuST} both want to predict the traffic density.
Therefore we will simply use the \term{Mean Squared Error (MSE)} over the whole test set to evaluate the accuracy.
Based on the fact that we analyze a fully distributed setting with windows for predictions, our formula for the \term{MSE} looks a bit different:

\begin{equation}
    MSE = \frac{1}{|y|} \sum\limits_{j=0}^{|N|} \sum\limits_{i=0}^{|D_{test}|} \sum\limits_{t=0}^{\frac{|D_{test}|}{w}}\big(y_{ijt}-\hat{y}_{ijt}\big)^2
\end{equation}

We first iterate over every node from $N$.
For each node, we are iterating over the test data $D_{test}$ of the node and finally also over the sliced windows of size $w$.
The squared distance is calculated for each value $\hat{y}_{ijt}$.
Finally, all error values are summed up and normalized.

Before data is processed by the network, every input is transformed by the instance norm.
This is a typical normalization, used to standardize time series data.
For a time series $x$ with mean $\overline{x}$ and standard deviation $\sigma (x)$ over the time axis the instance norm is calculated for each feature channel by

\begin{align}
    norm_{inst} = \frac{x-\overline{x}}{\sigma (x)}
\end{align}

Before the outputs of the models are analyzed, they are transformed back to the original data space.
With the \term{MSE} metric we have the option, to compare centralized approaches directly with decentralized ones.

\subsection{Model Variations}

This section contains a short overview over the different approaches compared in the evaluation.

\subsubsection{kNN}

\term{kNN} is used as a baseline centralized classifier, which uses no privacy.
It is accessing data from all nodes in the network at the same time.
In the setting of city traffic, it compares the test sequence of graphs with every 12 subsequent graphs in the history and uses the nearest prediction in euclidean space.
As graphs are averaged over the buckets, we fixed the amount of nearest graphs to $k=1$.
With this assumption,   the label of the closest graph can be returned.

\subsubsection{LabelProportionLocal}

This model is a simple version of our approach and also works decentralized.
One main difference to our full approach exists.
No data is exchanged between the nodes.
Therefore each node just uses its own data to predict the future.
Nevertheless, the model consists of one \term{LSTM} $\rightarrow$ \term{ReLU} $\rightarrow$ \term{Local Linear Layer} $\rightarrow$ \term{Dense-Layer} layer chain.
The part that is missing is the exchange of histogram information from the neighbors.

\subsubsection{LabelProportionToDense}

The \term{LabelProportionsToDense} model type uses our full approach as described in \autoref{sec:approach}.
This means, that $\varepsilon$-differentially private histogram data is exchanged to make use of spatial locality.
Data transfer is happening by peer-to-peer communication and without disclosing private information by using noise as described in the \term{Differential Privacy} subsection.

\subsection{Experiment Settings}

The evaluation will show how our approach performs against a centralized algorithm.
Additionally, we want to show the approximate difference in needed communication for training as well as the influence of the privacy parameter $\varepsilon$.
The resulting comparison provides a general direction to indicate whether the accuracy of our approach is good enough compared to the other approaches with no or less privacy guarantees.

Later on we will vary the hyperparameter $\varepsilon$ that controls how much noise is applied to the histograms.
We will compare $\varepsilon=0.1$, $\varepsilon=0.5$ and \term{no privacy gain} to give a meaningful evaluation of the privacy factor.
The general definition of \term{Differential Privacy} states that the higher the value of $\varepsilon$, the less privacy is guaranteed.
Therefore a value of $\varepsilon = 0.1$ gains a lot of privacy, but when $\varepsilon$ is set to $1$, no privacy is assured.

Other hyperparameters of the algorithms are set as follows:
The \term{kNN} uses $k=1$ for the number of nearest neighbors.
Deep Learning models, especially our approaches, are learned with the \term{ADAM optimizer} and with a fixed \term{learning rate} of $0.01$. The models are trained for $5$ epochs for the \term{Pems-Bay} and the \term{METR-LA} dataset, while the smaller \term{LuST} dataset is trained for $15$ epochs.

For the experiments, \term{Pems-Bay} and \term{METR-LA} datasets are used in the typical \term{train-test split} with distribution of $80$ percent train and $20$ percent test.
The \term{LuST} dataset however, is performed as a cross-validation with $5$ folds.
This is due to the fact that this dataset contains only information gathered in one day.
It can therefore happen, that the typical curve of rush hour in traffic density is never seen in training, but should be predicted in the test set.
Hence our decision to work with cross validation, to use each part of the dataset once for testing.

\subsection{Results}

Evaluation results are clustered into different sections.
First we will give an overview about the general performance dependent on the dataset and used algorithm.
Afterwards, we will explain concrete differences using specific extracts from the test data predictions.

\subsubsection{MSE}

An overview of the general test accuracy based on \term{MSE} is shown in \autoref{tab:test-mse-loss}.
\begin{table}[h]
    \centering
    \begin{tabular}{lllr}
\toprule
\tableHead{dataset} & \tableHead{model} & \tableHead{$\varepsilon$} & \tableHead{test loss} \\
\midrule
LuST & kNN & x &      0.470 \\
        & LabelProportionLocal &  x &      \textbf{0.377} \\
        & LabelProportionToDense &  x &      0.379 \\
        &                        &  0.5 &      0.381 \\
        &                        &  0.1 &      0.380 \\ \hline
METR-LA & kNN & x &      1.020 \\
        & LabelProportionLocal & x &      0.760 \\
        & LabelProportionToDense & x &      \textbf{0.480} \\
        &                        &  0.5 &      0.664 \\
        &                        &  0.1 &      0.730 \\ \hline
Pems-Bay & kNN &  x &      0.369 \\
        & LabelProportionLocal & x &      0.496 \\
        & LabelProportionToDense & x &      \textbf{0.305} \\
        &                        &  0.5 &      0.487 \\
        &                        &  0.1 &      0.542 \\
\bottomrule
\end{tabular}
    \caption{Overview of the MSE test losses. Each algorithm with changed $\varepsilon$ privacy parameter is tested on each dataset.}
    \label{tab:test-mse-loss}
\end{table}
\begin{figure*}[h]
    \centering
    \begin{tikzpicture}
    	\begin{axis}[
    			width=18cm,
    			height=6cm,
    			xmin=0,
    			xmax=10500,
    			ylabel=speed,
    			xlabel=time,
    			legend style={
    				draw=none, 
    				legend columns=4,
    				at={(0.5,-0.2)},
    				anchor=north,
    				column sep=0.5cm
    			}
    		]
    		\addplot[color=gray, solid, mark=none]
    		table[x=index, y=actual, col sep=semicolon] {pemsBayTestResults-full.csv};
    		\addplot[color=blue, dashed, mark=none, select coords between index={0}{3500}]
    		table[x=index, y=KNNCentralized, col sep=semicolon] {pemsBayTestResults-full.csv};
    		\addplot[color=orange, dashed, mark=none, select coords between index={3500}{7500}]
    		table[x=index, y=LabelProportionLocal, col sep=semicolon] {pemsBayTestResults-full.csv};
    		\addplot[color=black, dashed, mark=none, select coords between index={7500}{10500}]
    		table[x=index, y=LabelProportionToDense, col sep=semicolon] {pemsBayTestResults-full.csv};
    		\legend{actual, kNN, LabelProportionsLocal, LabelProportionsToDense};
    	\end{axis}
    \end{tikzpicture}
    \caption{Testset Prediction for one node in 5 minute prediction intervals on the Pems-Bay dataset. The solid gray line represents the ground truth, whereas the blue, orange and black dotted lines are the predictions of different algorithms. For clarity, only parts of the test set were plotted for each algorithm.}
    \label{fig:pemsbay_full}
\end{figure*}
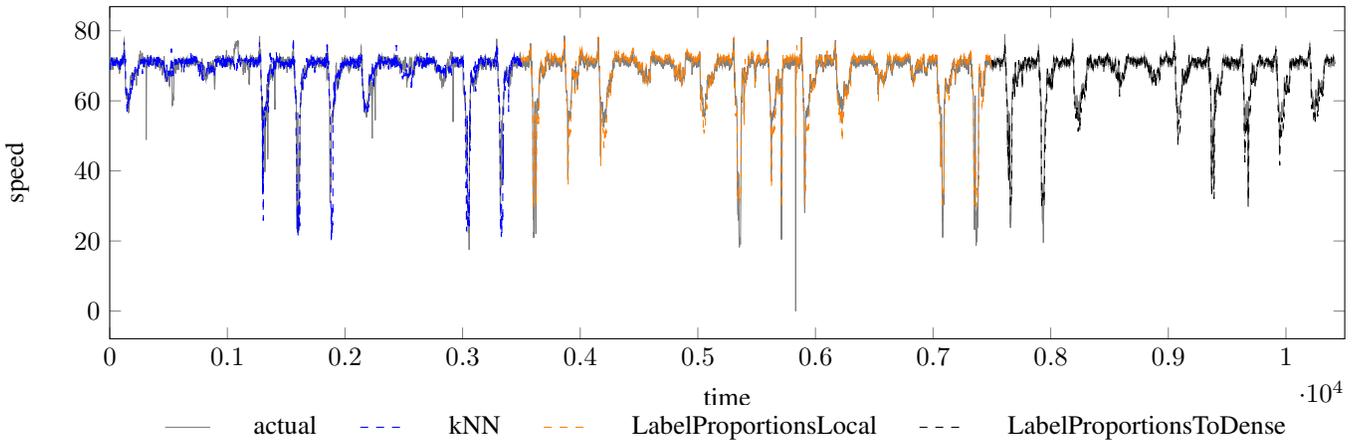
The first column denotes the used dataset whereas the second and third column denote the model and $\varepsilon$ privacy parameter.
When no privacy degree $\varepsilon$ can be specified, the value is shown as x.
It stands out that the metric ranges from $0.3$ to $1.02$, lower being better.
Obviously, the general performance is highly dependent on the datasets. 
For example, the \term{MSE} of the \term{LuST} dataset is around $0.38$ to $0.47$, whereas no value is below $0.48$ on the \term{METR-LA} dataset.

Looking at the figures, it is obvious that the \term{LabelProportionLocal} or \term{LabelProportionToDense} algorithm always achieves better results than the \term{kNN} as a centralized approach.
This is probably due to the fact that the integrated \term{LSTM} in our approach is better able to represent temporal components.
Another noticeable aspect is the increasing \term{MSE} value for increasing \term{privacy guarantees} by reducing $\varepsilon$.
This is also an expected degradation in terms of accuracy, since adding noise reduces the information content of the neighbors.
But it is interesting to see that the performance is getting worse compared to the \term{LabelProportionLocal} algorithm, which is not using additional neighbor information.
Only when using no noise, performance of \term{LabelProportionToDense} is the best on \term{METR-LA} and \term{Pems-Bay} in comparison to the other algorithms.

Based on these general results, one can say that our approach with sending \term{histograms} between direct nodes is improving the general prediction performance.
By using the \term{LSTM} as the central learning model for local data, we achieved to outperform the centralized \term{kNN} algorithm.
Using \term{$\varepsilon$-Differential Privacy} to ensure privacy of exchanged data, we have measured a significant increase of the \term{MSE} error, which results in worse performance, than when using no neighbor information.

\subsubsection{Prediction Curve Pems-Bay - Overview}

Because a single metric is not very meaningful, we plotted the predicted values of the different algorithms on the \term{Pems-Bay} dataset in \autoref{fig:pemsbay_full}.
The actual measured data is depicted by the solid grey line in the background.
In the foreground the three different algorithms are compared to each other.
This chart contains no privacy preserving approach with privacy parameter $\varepsilon$.
Predictions of the algorithms are split into sections on the test set, for better visibility.
Therefore, the predictions of the \term{kNN} as the blue line is only plotted for the first 3500 steps.
Following this, the prediction of the \term{LabelProportionLocal} has been plotted in orange up to time step 7500.
Finally, the predicted values of the \term{LabelProportionToDense} approach are shown in black.
As seen at the time scale we have predictions of approximately 10500 time steps which are equal to $10500 * 5 min = 52500 min = 875 h \approx 36$ days.

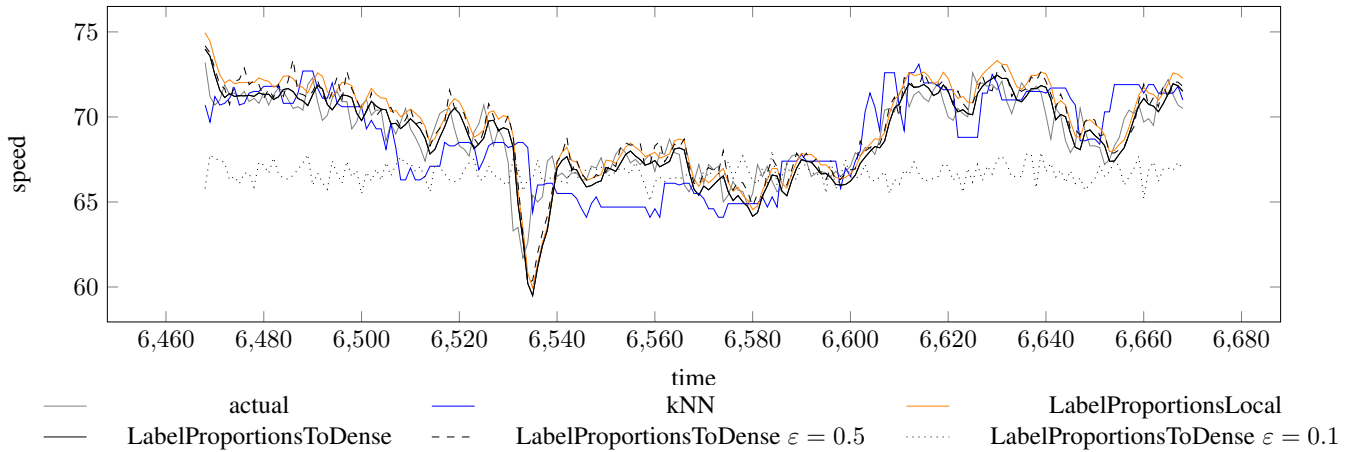
\begin{figure*}
    \centering
    \begin{tikzpicture}[scale=0.95]
    	\begin{axis}[
    			width=18cm,
    			height=6cm,
    			ylabel=speed,
    			xlabel=time,
    			legend style={
    				draw=none, 
    				legend columns=3,
    				at={(0.5,-0.2)},
    				anchor=north,
    				column sep=0.5cm
    			}
    		]
    		\addplot[color=gray, solid, mark=none]
    		table[x=index, y=actual, col sep=semicolon] {pemsBayTestResults-best.csv};
    		\addplot[color=blue, solid, mark=none]
    		table[x=index, y=KNNCentralized, col sep=semicolon] {pemsBayTestResults-best.csv};
    		\addplot[color=orange, solid, mark=none]
    		table[x=index, y=LabelProportionLocal, col sep=semicolon] {pemsBayTestResults-best.csv};

    		\addplot[color=black, solid, mark=none]
    		table[x=index, y=LabelProportionToDense, col sep=semicolon] {pemsBayTestResults-best.csv};
    		\addplot[color=black, dashed, mark=none]
    		table[x=index, y=LabelProportionToDense-0.5, col sep=semicolon] {pemsBayTestResults-best.csv};
    		\addplot[color=black, dotted, mark=none]
    		table[x=index, y=LabelProportionToDense-0.1, col sep=semicolon] {pemsBayTestResults-best.csv};
    		\addplot[color=black, solid, mark=none]
    		table[x=index, y=LabelProportionToDense, col sep=semicolon] {pemsBayTestResults-best.csv};
    		\legend{actual, kNN, LabelProportionsLocal, LabelProportionsToDense, LabelProportionsToDense $\varepsilon = 0.5$, LabelProportionsToDense $\varepsilon = 0.1$};
    	\end{axis}
    \end{tikzpicture}
    \caption{Testset slice of 200 time stamps. Prediction for one node in 5 minute prediction intervals on the PEMS-BAY dataset. The solid gray line represents the ground truth, whereas the blue, orange and black solid lines are the predictions of different algorithms. For clarity, only parts of the test set were plotted for each algorithm. When $\varepsilon$-Differential Privacy is applied, the lines are dashed or dotted.}
    \label{fig:pemsbay_200}
\end{figure*}

Therefore, only the general prediction shape can be seen, which is quite accurate.
At most times, the \term{kNN} and our approaches nearly met the actual prediction curve.
There are some exceptions, where the data cannot be fitted very well.
Especially peaks in the actual car speed measurements are not recognized by the approaches.
For example, the \term{kNN} does not predict the sharp drop in speed at around time step 150.
The same issue occurs, when looking at the \term{LabelProportionLocal} approach around time step 5800.
The only algorithm, that fits the curve nearly perfectly by viewing this in large scale is the \term{LabelProportionToDense} approach.

In this overview, the cycles where traffic speed is going down during rush hour are very good to see.
During the normal day, the density is oscillating around 70 which is the normal expected traffic speed curve.
This can be the reason why our approaches reach better prediction accuracy, because \term{time dependent} features are extracted by the LSTM.

\subsubsection{PredictionCurve Pems-Bay - Detailed}

In the plot over the entire test period, you can see tendencies, indicating the prediction accuracy.
For better details, we have cut out a section of 200 time steps and displayed it in \autoref{fig:pemsbay_200}.
This plot represents $200 * 5min = 1000min \approx 17h$, which is part of a day and night phase.
It can also be seen by assuming that rush-hour is visible in the time steps starting from 6535.
The night phase then begins around time step 6610, where a near constant speed is driven.
We chose the slice by analyzing the dataset for the area, where the \term{MSE} was the lowest.
Therefore, we chose the concrete slice from time frame 6468 to 6668.

The actual measurement values are visible in gray.
The kNN, as well as the \term{LabelProportionLocal} and \term{LabelProportionToDense} are shown solid in the same color as the previous chart.
Additionally there are two more variations plotted.
Those are our approaches (\term{LabelProportionToDense}), where noise is applied by $\varepsilon$-Differential Privacy.
Variations with noise are highlighted by dashed ($\varepsilon = 0.5$) or dotted ($\varepsilon = 0.1$) lines.

The figure indicates, it is clear that the prediction is not as accurate as indicated in \autoref{fig:pemsbay_full}.
Here, all little deviations in density prediction are getting noticed.

By focusing on the \term{kNN} it looks like it is mostly under predicting the real world measurements.
Only for steep drops, visible at around time step 6535, the prediction is not close to the real curve.
In general, the prediction of \term{kNN} looks somewhat like a step function. 
Therefore, all the little variations are not predicted well.

Compared to this, our approach without histogram transfer (\term{LabelProportionLocal}) fits the real world measurements quite well.
Especially the steep drop, which the \term{kNN} could not handle, is fitted well.
For most predictions it is just slightly above the real world data and at some peaks, like in time step 6590, it is shifted along the time axis.
At those points, the peak is predicted a bit later.
But in general, this prediction curve is quite close to the original measurements.

The only better approach than this is the \term{LabelProportionToDense} approach.
Variations between the prediction of the approach and the real data is almost not visible.
Only slight jumps of the original data, which are not relevant for the general traffic speed, are not predicted.
By looking at \autoref{fig:pemsbay_200}, this algorithm is approaching the best results from all tested ones.

However no privacy guarantee can be given for the histograms.
Therefore we added \term{differentially private} variations.
For $\varepsilon = 0.5$, the dashed curve shows the predictions.
Those predictions are also quite good and fit the real world data well.
Sometimes peaks are predicted with no real speed peak.
This can be seen at time step 6485.

For the \term{LabelProportionToDense} with $\varepsilon = 0.1$, the dotted line shows the predictions.
Those are much worse than all other algorithms.
When analyzing the curve, one can see that it is mostly a prediction around a traffic density of 66.
It varies in the prediction value, but not by much.
Therefore, it looks like the approach has learned to just predict the mean value.
The reason for this can be the applied noise to the histograms.
The noise could have resulted in nearly equally distributed histograms, so that no information is included.
When this happened, the model could also learn only the mean distribution, or mean value.
For this reason, it looks like a privacy factor of $\varepsilon = 0.1$ is too high to gain useful information from neighboring histograms that are afterwards normalized again over all neighbors.



\section{Conclusion}

In conclusion, the general approach adopted from \cite{DulacArnold2019:DeepMultiClassLLP} results in very good prediction accuracy on spatio-temporal data, as used in our evaluation with \term{Pems-Bay}, \term{METR-LA} and \term{LuST}.
We could show, that the local approach of using an \term{LSTM} combined with \term{ReLU} and \term{Dense-Layer} results in a very good prediction, because temporal information is extracted well by the \term{LSTM} model.
By adding information of the neighbors when integrating histograms, we could improve results by $0.28$ on \term{METR-LA} and $0.19$ on \term{Pems-Bay} as shown in \autoref{tab:test-mse-loss} for the \term{Pems-Bay} dataset.
However, adding \term{Differential Privacy} to the neighboring histograms, has a significant impact on the learning performance.
As shown by the evaluation, sometimes it is better to use no noisy neighbor data to train the model.

For the future, it is conceivable that either the second-degree neighbors will be included or that an attempt will be made to transfer \term{differentially private} data with less information loss or misinformation.

\FloatBarrier
\bibliographystyle{IEEEtran}
\bibliography{literature}

\begin{thebibliography}{10}
\providecommand{\url}[1]{#1}
\csname url@samestyle\endcsname
\providecommand{\newblock}{\relax}
\providecommand{\bibinfo}[2]{#2}
\providecommand{\BIBentrySTDinterwordspacing}{\spaceskip=0pt\relax}
\providecommand{\BIBentryALTinterwordstretchfactor}{4}
\providecommand{\BIBentryALTinterwordspacing}{\spaceskip=\fontdimen2\font plus
\BIBentryALTinterwordstretchfactor\fontdimen3\font minus
  \fontdimen4\font\relax}
\providecommand{\BIBforeignlanguage}[2]{{%
\expandafter\ifx\csname l@#1\endcsname\relax
\typeout{** WARNING: IEEEtran.bst: No hyphenation pattern has been}%
\typeout{** loaded for the language `#1'. Using the pattern for}%
\typeout{** the default language instead.}%
\else
\language=\csname l@#1\endcsname
\fi
#2}}
\providecommand{\BIBdecl}{\relax}
\BIBdecl

\bibitem{Mondschein2019}
\BIBentryALTinterwordspacing
C.~F. Mondschein and C.~Monda, \emph{The EU's General Data Protection
  Regulation (GDPR) in a Research Context}.\hskip 1em plus 0.5em minus
  0.4em\relax Cham: Springer International Publishing, 2019, pp. 55--71.
  [Online]. Available: \url{https://doi.org/10.1007/978-3-319-99713-1\_5}
\BIBentrySTDinterwordspacing

\bibitem{Sachweh2021:llp}
\BIBentryALTinterwordspacing
T.~Sachweh, D.~Boiar, and T.~Liebig, ``Differentially private learning from
  label proportions,'' in \emph{Machine Learning and Principles and Practice of
  Knowledge Discovery in Databases - International Workshops of {ECML} {PKDD}
  2021, Virtual Event, September 13-17, 2021, Proceedings, Part {I}}, ser.
  Communications in Computer and Information Science, M.~Kamp, I.~Koprinska,
  A.~Bibal, and Others, Eds., vol. 1524.\hskip 1em plus 0.5em minus 0.4em\relax
  Springer, 2021, pp. 119--127. [Online]. Available:
  \url{https://doi.org/10.1007/978-3-030-93736-2\_11}
\BIBentrySTDinterwordspacing

\bibitem{4470249}
D.~R. Musicant, J.~M. Christensen, and J.~F. Olson, ``Supervised learning by
  training on aggregate outputs,'' in \emph{Seventh IEEE International
  Conference on Data Mining (ICDM 2007)}, 2007, pp. 252--261.

\bibitem{stolpe2011learning}
M.~Stolpe and K.~Morik, ``Learning from label proportions by optimizing cluster
  model selection,'' in \emph{Joint European Conference on Machine Learning and
  Knowledge Discovery in Databases}.\hskip 1em plus 0.5em minus 0.4em\relax
  Springer, 2011, pp. 349--364.

\bibitem{Liebig2015:labelproportions}
\BIBentryALTinterwordspacing
T.~Liebig, M.~Stolpe, and K.~Morik, ``Distributed traffic flow prediction with
  label proportions: From in-network towards high performance computation with
  {MPI},'' in \emph{Proceedings of the 2nd International Workshop on Mining
  Urban Data co-located with 32nd International Conference on Machine Learning
  {(ICML} 2015), Lille, France, July 11th, 2015}, ser. {CEUR} Workshop
  Proceedings, I.~Katakis, F.~Schnitzler, T.~Liebig, D.~Gunopulos, K.~Morik,
  G.~L. Andrienko, and S.~Mannor, Eds., vol. 1392.\hskip 1em plus 0.5em minus
  0.4em\relax CEUR-WS.org, 2015, pp. 36--43. [Online]. Available:
  \url{http://ceur-ws.org/Vol-1392/paper-05.pdf}
\BIBentrySTDinterwordspacing

\bibitem{DulacArnold2019:DeepMultiClassLLP}
\BIBentryALTinterwordspacing
G.~Dulac{-}Arnold, N.~Zeghidour, M.~Cuturi, L.~Beyer, and J.~Vert, ``Deep
  multi-class learning from label proportions,'' \emph{CoRR}, vol.
  abs/1905.12909, 2019. [Online]. Available:
  \url{http://arxiv.org/abs/1905.12909}
\BIBentrySTDinterwordspacing

\bibitem{Yu2018:SpatioTempGCNNTraffic}
\BIBentryALTinterwordspacing
B.~Yu, H.~Yin, and Z.~Zhu, ``Spatio-temporal graph convolutional networks: {A}
  deep learning framework for traffic forecasting,'' in \emph{Proceedings of
  the Twenty-Seventh International Joint Conference on Artificial Intelligence,
  {IJCAI} 2018, July 13-19, 2018, Stockholm, Sweden}, J.~Lang, Ed.\hskip 1em
  plus 0.5em minus 0.4em\relax ijcai.org, 2018, pp. 3634--3640. [Online].
  Available: \url{https://doi.org/10.24963/ijcai.2018/505}
\BIBentrySTDinterwordspacing

\bibitem{rivest1978method}
R.~L. Rivest, A.~Shamir, and L.~Adleman, ``A method for obtaining digital
  signatures and public-key cryptosystems,'' \emph{Communications of the ACM},
  vol.~21, no.~2, pp. 120--126, 1978.

\bibitem{acar2018survey}
A.~Acar, H.~Aksu, A.~S. Uluagac, and M.~Conti, ``A survey on homomorphic
  encryption schemes: Theory and implementation,'' \emph{ACM Computing Surveys
  (Csur)}, vol.~51, no.~4, pp. 1--35, 2018.

\bibitem{dwork2008differential}
C.~Dwork, ``Differential privacy: A survey of results,'' in \emph{International
  conference on theory and applications of models of computation}.\hskip 1em
  plus 0.5em minus 0.4em\relax Springer, 2008, pp. 1--19.

\bibitem{dwork2014algorithmic}
C.~Dwork, A.~Roth \emph{et~al.}, ``The algorithmic foundations of differential
  privacy.'' \emph{Found. Trends Theor. Comput. Sci.}, vol.~9, no. 3-4, pp.
  211--407, 2014.

\bibitem{hochreiter1997long}
S.~Hochreiter and J.~Schmidhuber, ``Long short-term memory,'' \emph{Neural
  computation}, vol.~9, no.~8, pp. 1735--1780, 1997.

\bibitem{huang2020lsgcn}
R.~Huang, C.~Huang, Y.~Liu, G.~Dai, and W.~Kong, ``Lsgcn: Long short-term
  traffic prediction with graph convolutional networks.'' in \emph{IJCAI},
  2020, pp. 2355--2361.

\bibitem{manolios1994first}
P.~Manolios and R.~Fanelli, ``First-order recurrent neural networks and
  deterministic finite state automata,'' \emph{Neural Computation}, vol.~6,
  no.~6, pp. 1155--1173, 1994.

\bibitem{yang2019k}
L.~Yang, Q.~Yang, Y.~Li, and Y.~Feng, ``K-nearest neighbor model based
  short-term traffic flow prediction method,'' in \emph{2019 18th International
  Symposium on Distributed Computing and Applications for Business Engineering
  and Science (DCABES)}.\hskip 1em plus 0.5em minus 0.4em\relax IEEE, 2019, pp.
  27--30.

\bibitem{knnFirstApproach}
\BIBentryALTinterwordspacing
B.~W. Silverman and M.~C. Jones, ``E. fix and j.l. hodges (1951): An important
  contribution to nonparametric discriminant analysis and density estimation:
  Commentary on fix and hodges (1951),'' \emph{International Statistical Review
  / Revue Internationale de Statistique}, vol.~57, no.~3, pp. 233--238, 1989.
  [Online]. Available: \url{http://www.jstor.org/stable/1403796}
\BIBentrySTDinterwordspacing

\bibitem{Hochreiter97:LSTM}
\BIBentryALTinterwordspacing
S.~Hochreiter and J.~Schmidhuber, ``Long short-term memory,'' \emph{Neural
  Comput.}, vol.~9, no.~8, pp. 1735--1780, 1997. [Online]. Available:
  \url{https://doi.org/10.1162/neco.1997.9.8.1735}
\BIBentrySTDinterwordspacing

\bibitem{Guo:IMV2019}
\BIBentryALTinterwordspacing
T.~Guo, T.~Lin, and N.~Antulov{-}Fantulin, ``Exploring interpretable {LSTM}
  neural networks over multi-variable data,'' in \emph{Proceedings of the 36th
  International Conference on Machine Learning, {ICML} 2019, 9-15 June 2019,
  Long Beach, California, {USA}}, ser. Proceedings of Machine Learning
  Research, K.~Chaudhuri and R.~Salakhutdinov, Eds., vol.~97.\hskip 1em plus
  0.5em minus 0.4em\relax {PMLR}, 2019, pp. 2494--2504. [Online]. Available:
  \url{http://proceedings.mlr.press/v97/guo19b.html}
\BIBentrySTDinterwordspacing

\bibitem{NEURIPS2019_9015}
\BIBentryALTinterwordspacing
A.~Paszke, S.~Gross, F.~Massa, A.~Lerer, J.~Bradbury, G.~Chanan, T.~Killeen,
  Z.~Lin, N.~Gimelshein, L.~Antiga, A.~Desmaison, A.~Kopf, E.~Yang, Z.~DeVito,
  M.~Raison, A.~Tejani, S.~Chilamkurthy, B.~Steiner, L.~Fang, J.~Bai, and
  S.~Chintala, ``Pytorch: An imperative style, high-performance deep learning
  library,'' in \emph{Advances in Neural Information Processing Systems 32},
  H.~Wallach, H.~Larochelle, A.~Beygelzimer, F.~d\textquotesingle
  Alch\'{e}-Buc, E.~Fox, and R.~Garnett, Eds.\hskip 1em plus 0.5em minus
  0.4em\relax Curran Associates, Inc., 2019, pp. 8024--8035. [Online].
  Available:
  \url{http://papers.neurips.cc/paper/9015-pytorch-an-imperative-style-high-performance-deep-learning-library.pdf}
\BIBentrySTDinterwordspacing

\bibitem{Rozemberczki2021:TorchGeoTemporal}
\BIBentryALTinterwordspacing
B.~Rozemberczki, P.~Scherer, Y.~He, G.~Panagopoulos, A.~Riedel, M.~S.
  Astefanoaei, O.~Kiss, F.~B{\'{e}}res, G.~L{\'{o}}pez, N.~Collignon, and
  R.~Sarkar, ``Pytorch geometric temporal: Spatiotemporal signal processing
  with neural machine learning models,'' in \emph{{CIKM} '21: The 30th {ACM}
  International Conference on Information and Knowledge Management, Virtual
  Event, Queensland, Australia, November 1 - 5, 2021}, G.~Demartini, G.~Zuccon,
  J.~S. Culpepper, Z.~Huang, and H.~Tong, Eds.\hskip 1em plus 0.5em minus
  0.4em\relax {ACM}, 2021, pp. 4564--4573. [Online]. Available:
  \url{https://doi.org/10.1145/3459637.3482014}
\BIBentrySTDinterwordspacing

\bibitem{li2018dcrnn_traffic}
Y.~Li, R.~Yu, C.~Shahabi, and Y.~Liu, ``Diffusion convolutional recurrent
  neural network: Data-driven traffic forecasting,'' in \emph{International
  Conference on Learning Representations (ICLR '18)}, 2018.

\bibitem{jagadish2014big}
H.~V. Jagadish, J.~Gehrke, A.~Labrinidis, Y.~Papakonstantinou, J.~M. Patel,
  R.~Ramakrishnan, and C.~Shahabi, ``Big data and its technical challenges,''
  \emph{Communications of the ACM}, vol.~57, no.~7, pp. 86--94, 2014.

\bibitem{codeca2017:luxembourg}
L.~Codec\'a, R.~Frank, S.~Faye, and T.~Engel, ``{Luxembourg SUMO Traffic (LuST)
  Scenario: Traffic Demand Evaluation},'' \emph{IEEE Intelligent Transportation
  Systems Magazine}, vol.~9, no.~2, pp. 52--63, 2017.

\bibitem{7906642}
L.~Codeca, R.~Frank, S.~Faye, and T.~Engel, ``Luxembourg sumo traffic (lust)
  scenario: Traffic demand evaluation,'' \emph{IEEE Intelligent Transportation
  Systems Magazine}, vol.~9, no.~2, pp. 52--63, 2017.

\bibitem{Lopez2018:sumo}
\BIBentryALTinterwordspacing
P.~{\'{A}}. L{\'{o}}pez, M.~Behrisch, L.~Bieker{-}Walz, J.~Erdmann,
  Y.~Fl{\"{o}}tter{\"{o}}d, R.~Hilbrich, L.~L{\"{u}}cken, J.~Rummel, P.~Wagner,
  and E.~WieBner, ``Microscopic traffic simulation using {SUMO},'' in
  \emph{21st International Conference on Intelligent Transportation Systems,
  {ITSC} 2018, Maui, HI, USA, November 4-7, 2018}, W.~Zhang, A.~M. Bayen,
  J.~J.~S. Medina, and M.~J. Barth, Eds.\hskip 1em plus 0.5em minus 0.4em\relax
  {IEEE}, 2018, pp. 2575--2582. [Online]. Available:
  \url{https://doi.org/10.1109/ITSC.2018.8569938}
\BIBentrySTDinterwordspacing

\end{thebibliography}

\end{document}